\documentclass[preprint,12pt]{elsarticle}
\usepackage[margin=0.8in]{geometry}  
\usepackage{amsmath}  
\usepackage{amssymb}
\usepackage{multirow}
\usepackage{array}
\usepackage{caption}
\usepackage{natbib}
\usepackage{xcolor}
\usepackage{url}
\usepackage{hyperref}
\usepackage{rotating} 
\usepackage{booktabs}

\usepackage[linesnumbered,ruled,vlined]{algorithm2e}
\journal{}
\begin{document}

\begin{frontmatter}

\title{Trend-Aware Multi-Task Learning for Short-Term Energy Forecasting}

\author[inst1]{Fulong Yao\corref{cor1}}
\ead{yaof@cardiff.ac.uk}
\affiliation[inst1]{organization={School of Computer Science and Informatics},
            addressline={Cardiff University}, 
            city={Cardiff},
            postcode={CF10 3AT}, 
            country={UK}}

\author[inst2]{Wanqing Zhao}
\ead{wanqing.zhao@newcastle.ac.uk}
\affiliation[inst2]{organization={School of Computing},
            addressline={Newcastle University}, 
            city={Newcastle upon Tyne},
            postcode={NE4 5TG}, 
            country={UK}}
            
\author[inst3]{Chao Zheng}
\ead{c.zheng11@newcastle.ac.uk}
\affiliation[inst3]{organization={Business School},
            addressline={Newcastle University}, 
            city={Newcastle upon Tyne},
            postcode={NE1 4SE}, 
            country={UK}}

\author[inst4]{Xiaofei Han}
\ead{elxh@leeds.ac.uk}
\affiliation[inst4]{organization={School of Electronic and Electrical Engineering},
            addressline={University of Leeds}, 
            city={Leeds},
            postcode={LS2 9JT}, 
            country={UK}}
\cortext[cor1]{Corresponding author: Fulong Yao} 


\begin{abstract}

Short-term energy forecasting plays an important role in real-time operational decision-making, such as electricity market bidding and power system dispatch, where both numerical accuracy and correct directional signals are essential. However, most existing forecasting approaches formulate the problem purely as a regression task, limiting their ability to explicitly capture stepwise directional movements and trend consistency required for operational decisions. To address this limitation, this paper proposes a trend-aware multi-task forecasting framework that decomposes forecasting outputs into directional movements and deviation magnitudes relative to the latest observation, enabling both accurate numerical prediction and interpretable trend-aware outputs. The framework adopts a task-specific dual-stream architecture and explores key design choices for integrating trend and deviation information, including hard versus probabilistic trend representations, symmetric versus asymmetric deviation modelling, and parallel versus sequential conditioning strategies. To stabilize multi-task learning and reduce manual tuning, an uncertainty-aware task weighting scheme is incorporated to automatically balance directional classification, deviation regression, and final output prediction during training. Experimental results on real-world energy datasets demonstrate that the proposed framework achieves competitive numerical accuracy compared with state-of-the-art algorithms, while consistently improving trend prediction performance with moderate computational cost. This capability is particularly beneficial in short-term energy system management, where consistent directional forecasting can provide more reliable decision support for practical operational scenarios such as market bidding, resource scheduling, and risk-aware energy management. Code and data are available at: \href{https://anonymous.4open.science/r/ClReTS-3A62/README.md}{\textcolor{blue}{https://anonymous.4open.science/r/ClReTS-3A62/README.md}}

\end{abstract}

\end{frontmatter}

\section{INTRODUCTION} \label{sec:1}

Short-term forecasting is a critical component of operational decision making in modern energy systems, supporting a wide range of applications such as electricity market trading \cite{shen2024interpretable,beykirch2024value,nickelsen2025bayesian}, real-time dispatch \cite{qu2023segmented,tang2025risk}, reserve scheduling \cite{thran2025reserve,zheng2025data}, and intraday risk management \cite{miskiw2025coordinated,uniejewski2025smoothing}. In such settings, forecasts are typically required at short future horizons (ranging from the next minute to the next several hours) and are directly embedded into actionable decisions. Importantly, the practical value of a forecast in such operational contexts is not determined solely by numerical accuracy metrics such as the mean squared error (MSE), but also by whether the forecast correctly captures the directional movement and maintains trend consistency across future steps. For example, in electricity price forecasting, a numerically precise prediction with an incorrect direction may lead to systematically suboptimal bidding decisions, whereas a slightly less accurate forecast that correctly anticipates an upward or downward movement can be more valuable in practice. Similar considerations arise in risk-aware power system operation, where directionally consistent short-term signals enable proactive interventions and improved operational robustness \cite{tang2025risk}. These observations highlight a fundamental gap between conventional accuracy-oriented forecasting objectives and the decision-oriented requirements of short-term energy forecasting. In such settings, reliable directional signals are often as important as numerical accuracy, as they directly affect short-term operational decisions.


Over the past decade, significant advances in machine learning and deep learning have substantially improved forecasting performance in energy systems, particularly for electricity demand, renewable generation, and energy price prediction.
Early approaches employed convolutional neural networks (CNNs) to extract local temporal patterns \cite{wibawa2022time, durairaj2022convolutional}, as well as recurrent neural networks (RNNs), such as long short-term memory (LSTM) and gated recurrent unit (GRU), to capture sequential dependencies \cite{waqas2024critical, yunita2025performance}. Hybrid architectures like D-CNN-LSTM further combined convolutional and recurrent structures to model both local and short-range dynamics \cite{yao2022integrated}. More recently, Transformer-based architectures \cite{vaswani2017attention} have emerged as the dominant backbone for short energy forecasting. Representative variants include Informer \cite{zhou2021informer}, Autoformer \cite{wu2021autoformer}, FEDformer \cite{zhou2022fedformer}, TimesNet \cite{wu2023timesnet}, and iTransformer \cite{liu2023itransformer}, which improve efficiency and representation through sparse attention, decomposition mechanisms, frequency-domain modelling, or alternative modelling axes. These models have been widely adopted in energy forecasting tasks and have achieved state-of-the-art performance on various benchmarks \cite{yao2022integrated,amalou2022multivariate,fang2021novel,wang2022transformer,kim2024multi, antonesi2025hybrid,li2025transformer, feng2024energy, tang2025short}. Despite these advances, such deep forecasting models formulate short-term energy prediction as a pure regression problem, learning a direct mapping from recent observations to future values and optimizing point-wise losses such as MSE. While regression-centric formulations are effective in minimizing aggregate numerical errors, they provide limited explicit insight into step-wise directional movements and do not directly encourage trend consistency across forecast horizons. As a result, models that perform competitively in terms of value accuracy may still generate multi-step trajectories with unstable or inconsistent directional behavior, potentially limiting their reliability for real-world energy management and operational decision-making. This limitation becomes particularly pronounced in multi-step energy forecasting, where error accumulation and uncertainty can lead to inconsistent directional signals across future steps \cite{antonesi2025hybrid}. In such settings, short-term predictions are often used as decision signals rather than long-term projections. Therefore, accurately anticipating whether the target variable will rise or fall at each step, and maintaining directional consistency, can be as important as minimizing numerical errors.

There are several studies that attempt to jointly model directional movements and numerical values in short-term energy forecasting. For example, Wang et al. \cite{wang2019daily} proposed a daily-pattern–based electricity price forecasting framework, in which a classification module is first used to identify the expected price pattern of the next day, followed by a pattern-specific regression model to generate price forecasts. While effective in capturing coarse-grained regimes, such two-stage designs operate at a relatively coarse temporal resolution and do not provide step-wise directional supervision across forecast horizons. More generally, multi-task formulations that simultaneously predict a numerical target together with its up/down movement have been explored in related time-series domains, such as financial forecasting \cite{park2022stock}. These approaches demonstrate the potential benefits of combining classification and regression objectives, but typically rely on manually specified loss weights to balance different tasks. In practical energy forecasting scenarios, such hand-tuned weighting strategies are sensitive to differences in loss scale, noise characteristics, and convergence dynamics, which may lead to unstable training or suboptimal trade-offs between directional correctness and numerical accuracy. 

In parallel with accuracy improvements, the need for more transparent and structured outputs has become increasingly important for energy forecasting models, particularly in high-stakes energy applications \cite{baembitov2025interpretability}. Existing studies have primarily focused on improving transparency at the representation level, aiming to reveal how models internally process temporal patterns. For example, interval type-2 fuzzy neural networks \cite{yao2025self} provide rule-based inference mechanisms, yet their final outputs remain purely numerical and do not explicitly expose step-wise directional movements. More recent deep learning approaches introduce explicit signal decomposition to describe prediction patterns. Models such as DLinear and NLinear \cite{zeng2023transformers} demonstrate that simple trend–seasonal decomposition can outperform complex architectures, while Transformer variants including ETSformer \cite{woo2022etsformer}, Autoformer \cite{wu2021autoformer}, and FEDformer \cite{zhou2022fedformer} incorporate trend, seasonal, or frequency components into the model structure. Although effective in energy prediction, these methods operate primarily at the input or representation level and still produce regression-based outputs, without explicitly disentangling directional movements from deviation magnitudes at the prediction target. In contrast, decision-oriented energy forecasting often benefits from output representations that explicitly describe both the direction and magnitude of future changes. From this perspective, representation-level transparency does not necessarily provide direct decision-relevant information at the prediction output level. This observation motivates an output-level task reformulation, in which the prediction target is structured into direction and deviation components rather than only modifying internal representations. This formulation provides a more explicit description of step-wise changes and offers a structured output format that can be more directly aligned with short-term operational decision-making.

This paper proposes a multi-task learning framework for short-term energy forecasting that explicitly decomposes each future prediction into a directional movement and a deviation magnitude. By modelling trend direction via classification and deviation magnitude via regression, the proposed framework yields forecasts in the form of interpretable direction–magnitude pairs. This output-level decomposition not only improves numerical accuracy, but also provides decision-relevant insights into step-wise future dynamics. The main contributions of this paper can be summarized as follows:

i) We formulate short-term energy forecasting as an output-level trend–deviation learning problem, where each forecast step is represented by a directional movement and a deviation magnitude. This formulation provides a structured output format that makes the predicted direction and magnitude more transparent.

ii) We propose a multi-task learning framework that jointly optimizes trend classification and deviation regression for multi-step energy forecasting, together with a systematic analysis of key design choices, including hard versus probabilistic trend modelling, symmetric versus asymmetric deviation estimation, and parallel versus sequential conditioning strategies.

iii) To address scale mismatch and noise heterogeneity across energy prediction tasks, we incorporate an uncertainty-aware loss weighting strategy that adaptively balances classification, regression, and output prediction objectives during training, eliminating the need for manual tuning.

iv) Experiments on electricity price and power demand forecasting under multiple forecasting horizons demonstrate that the proposed framework achieves competitive numerical accuracy and consistently strong trend prediction accuracy, while maintaining moderate computational cost.

The remainder of this paper is organized as follows. Section \ref{sec:2} presents the proposed multi-task forecasting framework (CaReTS) and its dual-stream architecture. Section \ref{sec:3} introduces the specific CaReTS model variants. Section \ref{sec:4} reports experimental evaluations on the real-world hourly energy series, including comparisons with state-of-the-art methods and ablation studies. Finally, Section \ref{sec:5} briefly concludes the paper and outlines future research directions.

\section{CaReTS Framework} \label{sec:2}
This section introduces CaReTS, an multi-task learning framework for energy time series forecasting. Specifically, two types of CaReTS architectures are presented, each consisting of a classification branch that captures the stepwise trend of future values and a regression branch that estimates the corresponding deviations. Moreover, a multi-task loss formulation, together with an uncertainty-based loss weighting algorithm, is designed to jointly optimize three tasks including the output prediction, deviation estimation, and trend classification. 

\subsection{CaReTS Architecture}\label{sec:2.1}
\begin{figure*}[]
	\begin{center}
        \includegraphics[width=16.5cm, trim=5 5 5 5, clip]{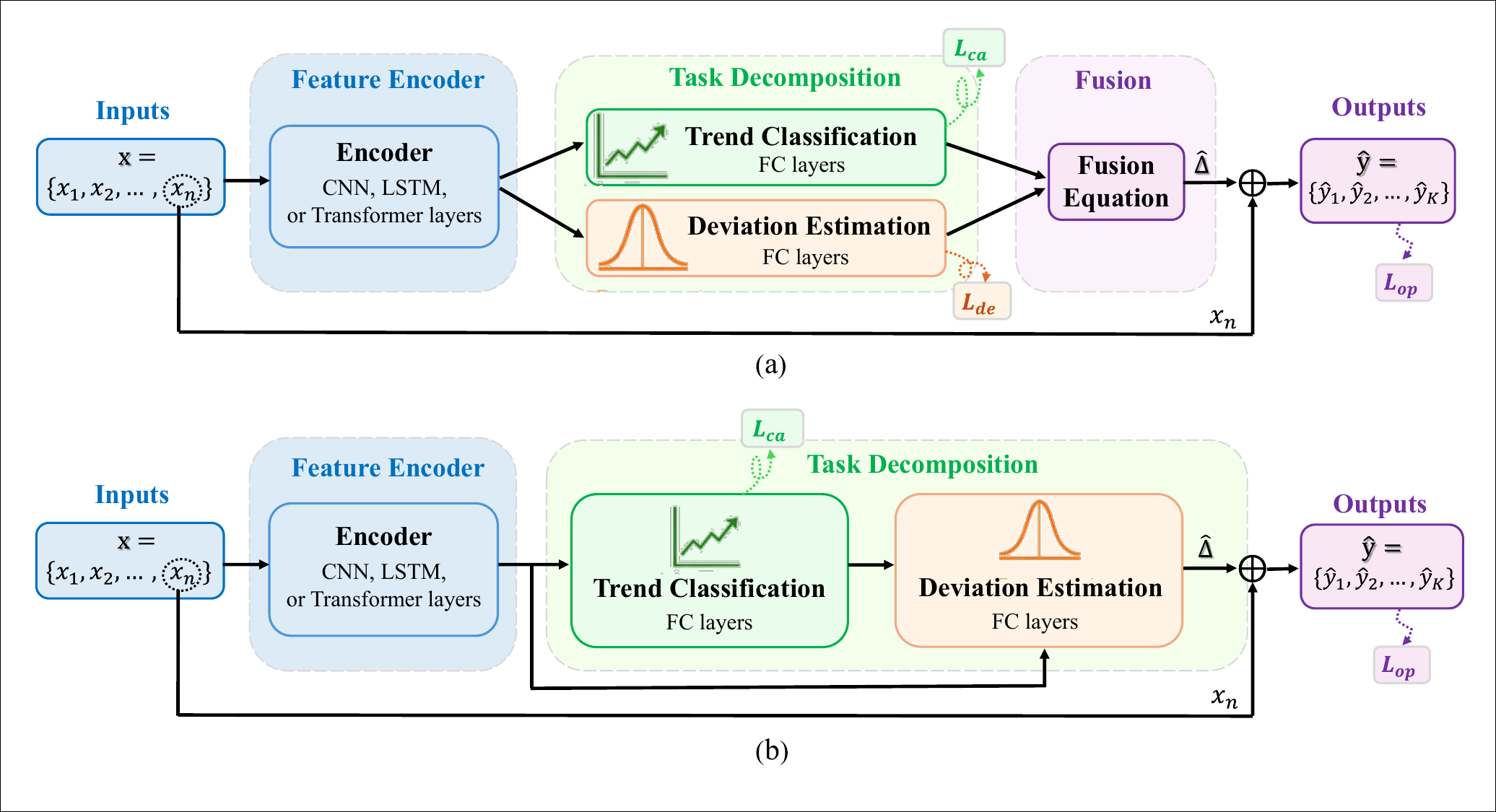}
           \captionsetup{skip=-3 pt}
		\caption{Two types of dual-stream CaReTS architectures} 
		\label{fig:1}
	\end{center}
\end{figure*}

Unlike traditional regression-based approaches that directly predict future values, this work designs two types of dual-stream CaReTS architectures that combine classification and regression tasks, as illustrated in Figure \ref{fig:1}. In both architectures, time series models such as CNNs, LSTMs, and Transformers are employed to encode temporal features from the input sequence $\textbf{x}  = \{x_1, x_2,…,x_n\}$. 
Here, $n$ denotes the total number of input variables, which may represent either a univariate or a multivariate time series. For instance, when predicting from the past three observations of two variables ($v_1$ and $v_2$), $\mathbf{x}$ is constructed as $\{v_1 (t-2), v_1 (t-1), v_1 (t), v_2 (t-2), v_2 (t-1), v_2 (t)\}$. Notably, the last entry $x_n$ (e.g., $v_2(t)$ in this example) must denote the most recent observation of the target variable.

The encoded temporal features are then processed through dual-stream pathways but differ in their fusion strategies. In architecture (a), these features are fed in parallel into two separate fully connected (FC) streams: a classification stream to model the stepwise trend (e.g., upward or downward) and a regression stream to estimate the corresponding deviations relative to the latest observation $x_n$.  The final prediction $\hat{\mathbf{y}}=\{\hat{y}_1,\hat{y}_2,\dots,\hat{y}_K\}=\{\hat{x}_{n+1},\hat{x}_{n+2},\dots,\hat{x}_{n+K}\}$  is obtained in a residual form that fuses the outputs of both streams with the last observation, i.e., the sum of $x_n$ and the predicted deviation $\hat{\boldsymbol{\Delta}} = \{\hat{\Delta}_1, \hat{\Delta}_2,\dots,\hat{\Delta}_K\}$. In contrast, architecture (b) adopts a sequential dual-stream design. The encoded temporal features are first processed by the classification stream to infer the trend. The resulting classification output is then concatenated with the original temporal features and passed into the regression stream for deviation estimation. Therefore, a separate fusion module is no longer required. Finally, similar to architecture (a), the final predictions $\hat{\textbf{y}}$ are produced by combining the predicted deviations $\hat{\boldsymbol{\Delta}}$ with the latest observation $x_n$.

The latest observation $x_n$ is used as the reference point because short-term operational forecasting is often interpreted relative to the current system state \cite{kong2025deep}. Compared with alternative references such as historical means, seasonal baselines, or moving averages, $x_n$ provides a local and immediately actionable anchor: the forecast can be interpreted as whether the target variable is expected to increase or decrease from its current value, and by how much. This choice is particularly suitable for short-horizon energy forecasting, where decisions such as bidding adjustment, dispatch correction, or short-term risk response rely on incremental changes relative to the current state \cite{vannoni2024integrated}. 


\subsection{Multi-Task Learning}\label{sec:2.2}
Building upon the dual-stream architectures introduced above, the CaReTS framework adopts a multi-task learning strategy to jointly model three interrelated tasks: trend classification, deviation estimation, and output prediction. This design is intended to improve forecasting accuracy while providing a structured separation between trend and magnitude components. In the context of energy forecasting, these tasks correspond to capturing directional signals (e.g., price increase or load variation) and their associated magnitudes, which are directly relevant for operational decision-making.

In architecture (a), all three tasks are learned in parallel. The overall loss function $L_{(a)}$ is formulated as:
\begin{equation} \label{eq:1}
        \small
		L_{(a)}= \alpha_{ca}L_{ca} + \alpha_{de}L_{de} + \alpha_{op}L_{op}
\end{equation}
where $\alpha_{ca},\alpha_{de}$, and $\alpha_{op}$ are the balancing weights of each tasks; $L_{ca}$ corresponds to the trend classification loss, evaluating the correctness of predicted trend (e.g., upward or downward movement) across multiple future steps. $L_{de}$ represents the deviation estimation loss, which supervises the intermediate magnitude component relative to $x_n$, while $L_{op}$ denotes the output prediction loss, which supervises the final reconstructed forecast after combining the trend and deviation components. These two losses play different roles: $L_{de}$ encourages the deviation branch to learn a meaningful magnitude representation, whereas $L_{op}$ ensures that the fused output remains numerically accurate. Therefore, $L_{de}$ acts as component-level supervision, while $L_{op}$ acts as prediction-level supervision. A detailed formulation of each loss will be presented in Section \ref{sec:3}. 

In contrast, architecture (b) simplifies the learning objective to two tasks, as it does not use an explicit fusion module. Here, the deviation estimation and output prediction are effectively combined into a single regression task, resulting in the following loss function:
\begin{equation} \label{eq:2}
        \small
		L_{(b)}= \alpha_{ca}L_{ca} + \alpha_{op}L_{op}
\end{equation}
By structuring the prediction process into distinct but related tasks, the CaReTS framework facilitates more transparent forecasting. It explicitly models how the trend influences the predicted output, thereby offering valuable insights for multi-step time series prediction.

Optimizing the three interrelated tasks simultaneously poses significant challenges, particularly due to discrepancies in loss scales, convergence dynamics, noise levels, and potential conflicts between task objectives. To address these issues, an uncertainty-based loss weighting algorithm \citep{kendall2018multi} is incorporated to adaptively adjust the contribution of each task during training. As defined in (\ref{eq:3}), each task's weight is modelled as the inverse of its predicted variance, reflecting the principle that tasks with higher uncertainty should contribute less to the overall loss. As a result, $\alpha_{ca},\alpha_{de}$, and $\alpha_{op}$ are not treated as static hyperparameters but are instead parameterized through learnable variables that capture the relative confidence of the model in each task. This formulation enables the model to focus on more informative and reliable tasks throughout the optimization process, thereby improving training stability and predictive performance.
\begin{equation} \label{eq:3}
        \small
	\begin{split}
		\alpha_i = \frac{1}{2\sigma_i^2}, \ \ \ \ i\in\{ca,de,op\}
	\end{split}
\end{equation}
where $\sigma_{ca}^2, \sigma_{de}^2$, and $\sigma_{op}^2$ represent the predicted variance (uncertainty) for each task.

In our implementation, the uncertainty-based task weights are modelled through their logarithmic counterparts ($\log \sigma_i^2$) to improve numerical stability and allow unconstrained gradient-based optimization. Specifically, each ($\log \sigma_i^2$) is treated as a learnable parameter, and the corresponding task weight is derived via exponential transformation. Accordingly, the overall loss function for architecture (a) and (b) are reformulated as: 
\begin{equation} \label{eq:4}
         \small
		L_{(a)} = \sum_{i \in \{\mathrm{ca}, \mathrm{de}, \mathrm{op}\}} 
\left( \frac{1}{2} e^{-\log\sigma_i^2} L_i + \frac{1}{2} \log\sigma_i^2 \right)
\end{equation}
\begin{equation} \label{eq:5}
         \small
		L_{(b)} = \sum_{i \in \{\mathrm{ca}, \mathrm{op}\}} 
\left( \frac{1}{2} e^{-\log\sigma_i^2} L_i + \frac{1}{2} \log\sigma_i^2 \right)
\end{equation}

This formulation integrates adaptive loss weighting with uncertainty regularization, allowing the model to adjust the relative importance of each subtask during training. It provides an alternative to manually fixed task weights and helps reduce sensitivity to differences in task scales. In this work, its practical effect is evaluated by comparing adaptive weighting with fixed-weight multi-task learning and single-task learning in Section \ref{sec:4.4}.

It should be noted that two stabilization strategies are used here to prevent pathological solutions (e.g., the model assigning arbitrarily large uncertainty to minimize its contribution). On the one hand, each log-variance parameter is softly regularized by an additional penalty term added to the total loss. On the other hand, during training, the log-variance values are constrained within a bounded range $[-10, 10]$ via clamping. These mechanisms jointly help stabilize uncertainty learning and avoid degenerate minima.

\section{CaReTS Models}\label{sec:3}

This section presents the design of the proposed approaches. To ensure broad applicability rather than introducing novel feature extractors, we adopt three mainstream temporal modelling algorithms, CNNs, LSTMs, and Transformers, as interchangeable encoders for extracting sequential features from the input time series.
These encoders are selected to cover convolutional, recurrent, and attention-based modelling mechanisms, while their detailed architectures are not the focus of this work and are therefore described in Section \ref{sec:4.1}. The primary focus of this section is the design of the CaReTS models. This formulation is motivated by the need to jointly model directional changes and magnitude variations in short-term energy forecasting tasks \cite{wang2019daily}. Building upon the dual-stream CaReTS architectures introduced in Section \ref{sec:2.1}, four specific model variants (CaReTS1, CaReTS2, CaReTS3, and CaReTS4) are developed to explore different strategies for combining classification-based trend modelling with regression-based deviation estimation, as illustrated in Table \ref{tab:1}. Specifically, CaReTS1–CaReTS3 adopt architecture (a), where the two streams operate in parallel; and CaReTS4 adopts architecture (b), where the trend prediction precedes and conditions the deviation estimation. Each model can be paired with any of the three temporal encoders (CNNs, LSTMs, and Transformers). 

\begin{table*}[htbp]
\centering
\caption{Comparison of CaReTS1--4 models}
\label{tab:1}
\resizebox{\textwidth}{!}{%
\begin{tabular}{ccp{2.85cm}p{3.5cm}p{3.8cm}p{3.7cm}}
\hline
\textbf{Model} & \textbf{Arch.} & \textbf{Trend} & \textbf{Deviation} & \textbf{Fusion} & \textbf{Loss} \\
\hline
\textbf{CaReTS1} 
& (a) 
& Binary label $\hat{d}^{(k)} \in \{+1,-1\}$ 
& Non-negative deviation $\hat{\delta}^{(k)}$ 
& $\hat{y}^{(k)} = x_n + \hat{d}^{(k)} \cdot \hat{\delta}^{(k)}$ 
& $L_{(a)}=L_{\mathrm{ca}} + L_{\mathrm{de}} + L_{\mathrm{op}}$ Eq.~(\ref{eq:9}), (\ref{eq:10}), (\ref{eq:11}) \\
\textbf{CaReTS2} 
& (a) 
& Binary label $\hat{d}^{(k)} \in \{+1,-1\}$ 
& Non-negative deviations $(\hat{\delta}^{(k)}_{\mathrm{up}}, \hat{\delta}^{(k)}_{\mathrm{down}})$
& If up: $\hat{y}^{(k)} = x_n + \hat{\delta}^{(k)}_{\mathrm{up}}$, else: $\hat{y}^{(k)} = x_n - \hat{\delta}^{(k)}_{\mathrm{down}}$
& $L_{(a)}=L_{\mathrm{ca}} + L_{\mathrm{de}} + L_{\mathrm{op}}$ Eq.~(\ref{eq:9}), (\ref{eq:13}), (\ref{eq:11}) \\
\textbf{CaReTS3} 
& (a) 
& Probabilities $(p^{(k)}_{\mathrm{up}}, p^{(k)}_{\mathrm{down}})$
& Non-negative deviations $(\hat{\delta}^{(k)}_{\mathrm{up}}, \hat{\delta}^{(k)}_{\mathrm{down}})$
& $\hat{y} = x_n + p^{(k)}_{\mathrm{up}}\hat{\delta}^{(k)}_{\mathrm{up}} - p^{(k)}_{\mathrm{down}}\hat{\delta}^{(k)}_{\mathrm{down}}$
& $L_{(a)}=L_{\mathrm{ca}} + L_{\mathrm{de}} + L_{\mathrm{op}}$ Eq.~(\ref{eq:16}), (\ref{eq:13}), (\ref{eq:11}) \\
\textbf{CaReTS4} 
& (b) 
& Probabilities $p^{(k)}$
& Signed deviation $\hat{\delta}^{(k)}$
& $\hat{y} = x_n + \hat{\delta}^{(k)}$
& $L_{(b)}=L_{\mathrm{ca}} + L_{\mathrm{op}}$ Eq.~(\ref{eq:18}), (\ref{eq:19}) \\
\hline
\end{tabular}
} 
\end{table*}

\subsection{CaReTS1}\label{sec:3.1}

This variant follows architecture (a) with two parallel fully connected (FC) streams. For each forecast step $k$, the trend branch predicts a binary class label 
$\hat{d}^{(k)} \in \{+1,-1\}$, where $+1$ denotes an upward trend and $-1$ represents a downward trend. In implementation, a single logit $z^{(k)}$ is predicted and transformed into a probability $p^{(k)} \in (0,1)$ via the sigmoid function:
\begin{equation} \label{eq:6}
        \small
    \hat{d}^{(k)} =
    \begin{cases}
    +1, & \text{if } p^{(k)} \geq 0.5, \\
    -1, & \text{if } p^{(k)} < 0.5,
    \end{cases}
\end{equation}
where
\begin{equation} \label{eq:7}
        \small
        p^{(k)} = \frac{1}{1 + e^{-z^{(k)}}}
\end{equation}
Meanwhile, the deviation branch predicts a single non-negative $\hat{\delta}^{(k)} \geq 0$, representing the absolute magnitude of change from the latest observation $x_n$, independent of direction.  

Finally, the forecast $\hat{y}^{(k)}$ is obtained by combining the predicted trend direction and magnitude:
\begin{equation} \label{eq:8}
    \small
    \hat{y}^{(k)} = x_n + \hat{d}^{(k)} \, \hat{\delta}^{(k)}
\end{equation}
The detailed loss of classification, deviation regression, and output prediction are defined as:
\begin{equation} \label{eq:9}
     \small
L_{\mathrm{ca}} = \frac{1}{K} \sum_{k=1}^K \mathrm{BCE}\left(p^{(k)}, t^{(k)}\right)
\end{equation}
\begin{equation} \label{eq:10}
     \small
L_{\mathrm{de}} = \frac{1}{K} \sum_{k=1}^K \mathrm{MSE}\left(\hat{\delta}^{(k)}, \delta^{(k)}\right)
\end{equation}
\begin{equation} \label{eq:11}
     \small
L_{\mathrm{op}} = \frac{1}{K} \sum_{k=1}^K \mathrm{MSE}\left(\hat{y}^{(k)}, y^{(k)}\right)
\end{equation}
where $t^{(k)} \in \{0, 1\}$ is the ground-truth trend label ($1$ for upward trend, $0$ for downward trend), $\delta^{(k)} = \left| y^{(k)} - x_n \right|$ is the true absolute deviation, $\mathrm{MSE}(\cdot,\cdot)$ denotes mean squared error, and $ \mathrm{BCE}\left(p^{(k)}, t^{(k)}\right) = -\left[t^{(k)} \log p^{(k)} + \left(1-t^{(k)}\right) \log \left(1-p^{(k)}\right)\right] $ denotes the binary cross-entropy loss. Note that the ground-truth trend label $t^{(k)}$ is used for BCE loss, which corresponds to $\hat{d}^{(k)} = +1$ when $t^{(k)}=1$ and $\hat{d}^{(k)} = -1$ when $t^{(k)}=0$.
CaReTS1’s simple architecture enables efficient parallel learning of trend movement and deviation magnitude, making the model both tractable and interpretable. Nevertheless, applying a uniform deviation magnitude across either directions in the trend means that any misclassification of trend inevitably results in forecast errors, with no capacity to capture direction-specific variations.

\subsection{CaReTS2}\label{sec:3.2}

CaReTS2 also adopts architecture (a) and retains the binary trend classifier from CaReTS1, but addresses one of CaReTS1’s limitations: the inability to differentiate magnitude patterns between upward and downward movements. Specifically, CaReTS2 replaces the single deviation output with direction-specific deviations, allowing the model to learn separate regression functions for positive and negative trends. This provides greater flexibility in capturing asymmetric dynamics or time series behaviors.  
Identical to CaReTS1, CaReTS2 outputs a binary trend label $\hat{d}^{(k)}$ as defined in (\ref{eq:6}). In contrast, its deviation branch produces two non-negative, direction-specific estimates: $\hat{\delta}_{\mathrm{up}}^{(k)}$ for upward movements and $\hat{\delta}_{\mathrm{down}}^{(k)}$ for downward movements. The final forecast then combines the predicted direction with the corresponding deviation, giving:
\begin{equation} \label{eq:12}
    \small
    \hat{y}^{(k)} =
    \begin{cases}
    x_n + \hat{\delta}_{\mathrm{up}}^{(k)}, & \text{if } \hat{d}^{(k)} = +1, \\
    x_n - \hat{\delta}_{\mathrm{down}}^{(k)}, & \text{if } \hat{d}^{(k)} = -1.
    \end{cases}
\end{equation}
The loss function retains the classification term $L_{\mathrm{ca}}$ and output prediction term $L_{\mathrm{op}}$ from CaReTS1. The deviation loss $L_{\mathrm{de}}$, however, is computed using the deviation estimate corresponding to the ground-truth trend direction:
\begin{equation} \label{eq:13}
     \small
    L_{\mathrm{de}} = \frac{1}{K} \sum_{k=1}^K \Big[ t^{(k)} \mathrm{MSE}\big(\hat{\delta}_{\mathrm{up}}^{(k)}, \delta_{\mathrm{up}}^{(k)}\big) 
    + \big(1 - t^{(k)}\big)  \mathrm{MSE}\big(\hat{\delta}_{\mathrm{down}}^{(k)}, \delta_{\mathrm{down}}^{(k)}\big) \Big]
\end{equation}
where $\delta_{\mathrm{up}}^{(k)} = \max\big(y^{(k)} - x_n, \, 0\big)
\quad$ and $\quad
\delta_{\mathrm{down}}^{(k)} = \max\big(x_n - y^{(k)}, \, 0\big)$
are the true upward and downward deviations, respectively.

\subsection{CaReTS3}\label{sec:3.3}

Similarly, CaReTS3 is based on architecture (a) and adopts the same deviation branch as CaReTS2, producing two separate non-negative estimates: $\hat{\delta}_{\mathrm{up}}^{(k)}$ for upward deviations and $\hat{\delta}_{\mathrm{down}}^{(k)}$ for downward deviations. 
The key change of CaReTS3 lies in the soft probabilistic trend modelling. Instead of producing a hard binary decision, the trend branch first generates a pair of logits $(z_{\mathrm{up}}^{(k)}, z_{\mathrm{down}}^{(k)})$, which are then transformed via a softmax into the output probabilities $(p_{\mathrm{up}}^{(k)}, p_{\mathrm{down}}^{(k)})$:
\begin{equation} \label{eq:14}
     \small
p_{\mathrm{up}}^{(k)} = \frac{e^{z_{\mathrm{up}}^{(k)}}}{e^{z_{\mathrm{up}}^{(k)}} + e^{z_{\mathrm{down}}^{(k)}}}, 
\quad
p_{\mathrm{down}}^{(k)} = 1 - p_{\mathrm{up}}^{(k)}
\end{equation}

Unlike selecting a single deviation value based on a hard sign decision, CaReTS3 fuses the two deviation predictions in a soft-weighted manner:
\begin{equation} \label{eq:15}
    \small
\hat{y}^{(k)} = x_n 
+ p_{\mathrm{up}}^{(k)}  \hat{\delta}_{\mathrm{up}}^{(k)} 
- p_{\mathrm{down}}^{(k)}  \hat{\delta}_{\mathrm{down}}^{(k)}
\end{equation}
This formulation allows both deviation predictions to contribute proportionally to the final forecast, enabling smoother transitions between upward and downward trends and potentially improving robustness when the trend movement is uncertain.

The loss design is same as CaReTS2 in terms of the deviation loss $L_{\mathrm{de}}$ and the output prediction loss $L_{\mathrm{op}}$,  
but the classification loss $L_{\mathrm{ca}}$ is redefined to handle probabilistic outputs:
\begin{equation} \label{eq:16}
     \small
L_{\mathrm{ca}} = \frac{1}{K} \sum_{k=1}^K 
\mathrm{CE}(\mathbf{p}^{(k)}, \mathbf{t}^{(k)})
\end{equation}
where $\mathbf{p}^{(k)} = \big(p_{\mathrm{up}}^{(k)}, \, p_{\mathrm{down}}^{(k)}\big), 
\quad 
\mathbf{t}^{(k)} = \big(t_{\mathrm{up}}^{(k)}, \, t_{\mathrm{down}}^{(k)}\big),
$
with $t_{\mathrm{up}}^{(k)} \in \{0,1\}$, $t_{\mathrm{down}}^{(k)} \in \{0,1\}$  
denoting the ground truth vector (1 for upward trend, 0 for downward trend), and the categorical cross-entropy is defined as: $\mathrm{CE}(\mathbf{p}^{(k)}, \mathbf{t}^{(k)}) 
= -\Big[ t_{\mathrm{up}}^{(k)} \log p_{\mathrm{up}}^{(k)} 
+ t_{\mathrm{down}}^{(k)} \log p_{\mathrm{down}}^{(k)} \Big].$

\subsection{CaReTS4}\label{sec:3.4}

CaReTS4 adopts architecture (b) and represents a sequential dual-stream approach, where the trend prediction stage precedes and conditions the deviation estimation stage. For each forecast step $k$, the model outputs the trend probability $p^{(k)}$ using a softmax-based classifier (similar to the trend branch of CaReTS1 and CaReTS2). Then, the predicted trend probabilities are concatenated with the temporal feature vector $\mathbf{h} \in \mathbb{R}^d$ extracted by the encoder, i.e., $
\mathbf{h}' = \big[ \mathbf{h}, \, \mathbf{p} \big]$.  
This operation allows the subsequent regression branch to condition its deviation estimation on the predicted trend context. Compared with the parallel design, where trend and deviation are inferred independently from shared temporal features, the sequential design assumes that directional information can serve as an additional explanatory signal for magnitude estimation. This strategy may be preferable when the magnitude of future change depends strongly on the predicted direction, such as during abrupt upward or downward movements. Using the fused feature vector $\mathbf{h}'$ as input, the model predicts a single signed deviation $\hat{\boldsymbol{\delta}}$,  
which may be positive or negative. This design differs fundamentally from CaReTS1--CaReTS3, where deviations were constrained to be non-negative and combined with a separate trend sign. 

The final forecast is obtained as:
\begin{equation} \label{eq:17}
    \small
\hat{y}^{(k)} = x_n + \hat{\delta}^{(k)}
\end{equation}
In contrast to earlier variants, CaReTS4 does not include a separate deviation loss $L_{\mathrm{de}}$. Instead, the training jointly optimizes two objectives:
\begin{equation} \label{eq:18}
     \small
L_{\mathrm{ca}} = \frac{1}{K} \sum_{k=1}^K \mathrm{CE}\big(p^{(k)}, t^{(k)}\big)
\end{equation}
\begin{equation} \label{eq:19}
     \small
L_{\mathrm{op}} = \frac{1}{K} \sum_{k=1}^K \mathrm{MSE}\big(\hat{y}^{(k)}, y^{(k)}\big)
\end{equation}

The above four variants can be interpreted as controlled instantiations of three underlying modelling dimensions, rather than as unrelated architectures. The first dimension is hard versus probabilistic trend representation, which determines whether the directional branch produces a discrete sign or a soft directional confidence. This distinction is reflected in models such as CaReTS1 and CaReTS2 (hard labels) versus CaReTS3 and CaReTS4 (probabilistic outputs). The second dimension is shared versus direction-specific deviation estimation, which determines whether upward and downward movements share the same magnitude model or are modelled separately. This is captured by comparing models with a single deviation output (e.g., CaReTS1) against those with direction-specific deviations (e.g., CaReTS2 and CaReTS3). The third dimension is parallel versus sequential conditioning, which determines whether trend and deviation are estimated independently from shared features or whether the predicted trend information is used to condition deviation estimation. This corresponds to the difference between the parallel architectures (CaReTS1–CaReTS3) and the sequential architecture (CaReTS4). These variants are intended to provide a structured comparison of different direction--magnitude interaction mechanisms in short-term energy forecasting.

\section{Experiments and Evaluation}\label{sec:4}

\begin{figure}[]
    \centering
    \setlength{\abovecaptionskip}{-3 pt} 
    \begin{minipage}{0.48\textwidth}
        \centering
        \includegraphics[width=\textwidth]{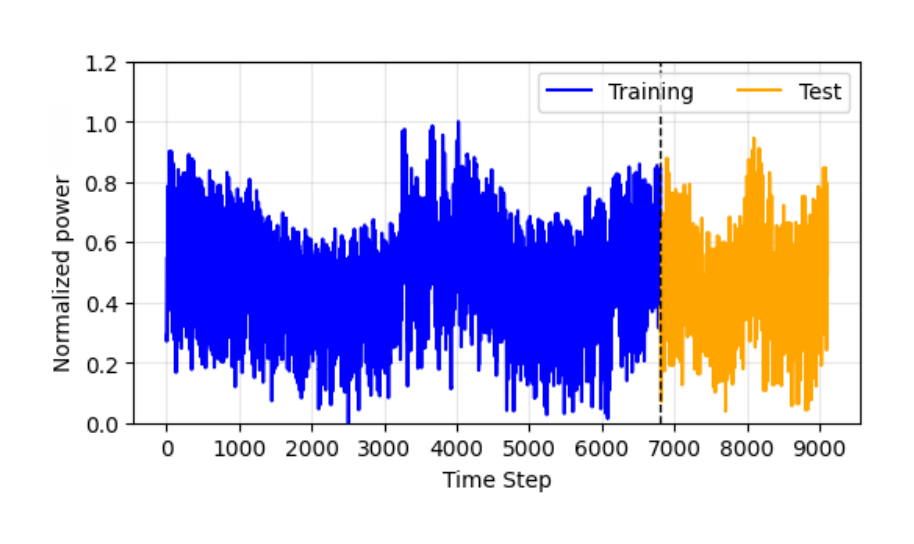}
        \caption{Illustration of unmet power}
        \label{fig:A.3-1}
    \end{minipage}\hfill
    \begin{minipage}{0.48\textwidth}
        \centering
        \includegraphics[width=\textwidth]{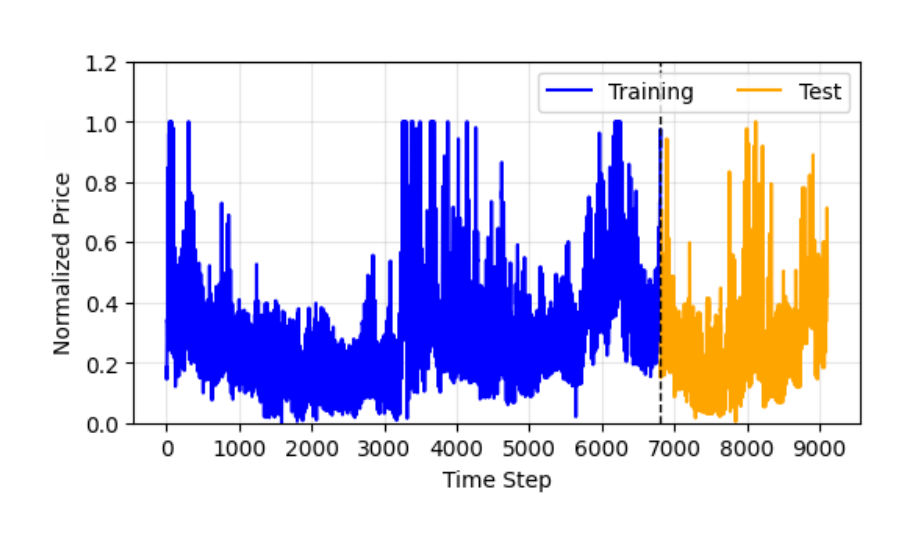}
        \caption{Illustration of electricity price}
        \label{fig:A.3-2}
    \end{minipage}
\end{figure}

In this section, we present a comprehensive experimental evaluation of the proposed CaReTS framework, including comparisons with baselines and state-of-the-art (SOTA) methods as well as ablation analyses. The evaluation was performed on two real-world time-series forecasting tasks derived from a U.S. district energy system \citep{yao2025self, yao2025holistic}: (i) electricity price forecasting and (ii) import/export power demand forecasting (i.e., unmet power). Both datasets span one year and consist of 8,784 hourly observations. To account for seasonal variability, the training and test sets were constructed by selecting sensor measurements from the first 21 days of each month as training data, while the remaining days were reserved for testing. This resulted in 6,048 training samples and 2,736 test samples for each task. As illustrated in Figures \ref{fig:A.3-1} and \ref{fig:A.3-2}, the two time series exhibit markedly different temporal patterns, providing a diverse testbed for assessing the robustness and generalization capability of forecasting models under varying conditions. A detailed description of the dataset can be found in \cite{yao2025self}.

\subsection{Experimental Settings}\label{sec:4.1}

\begin{figure}[htbp]
	\begin{center}
        \includegraphics[width=15cm, trim=5 5 5 5, clip]{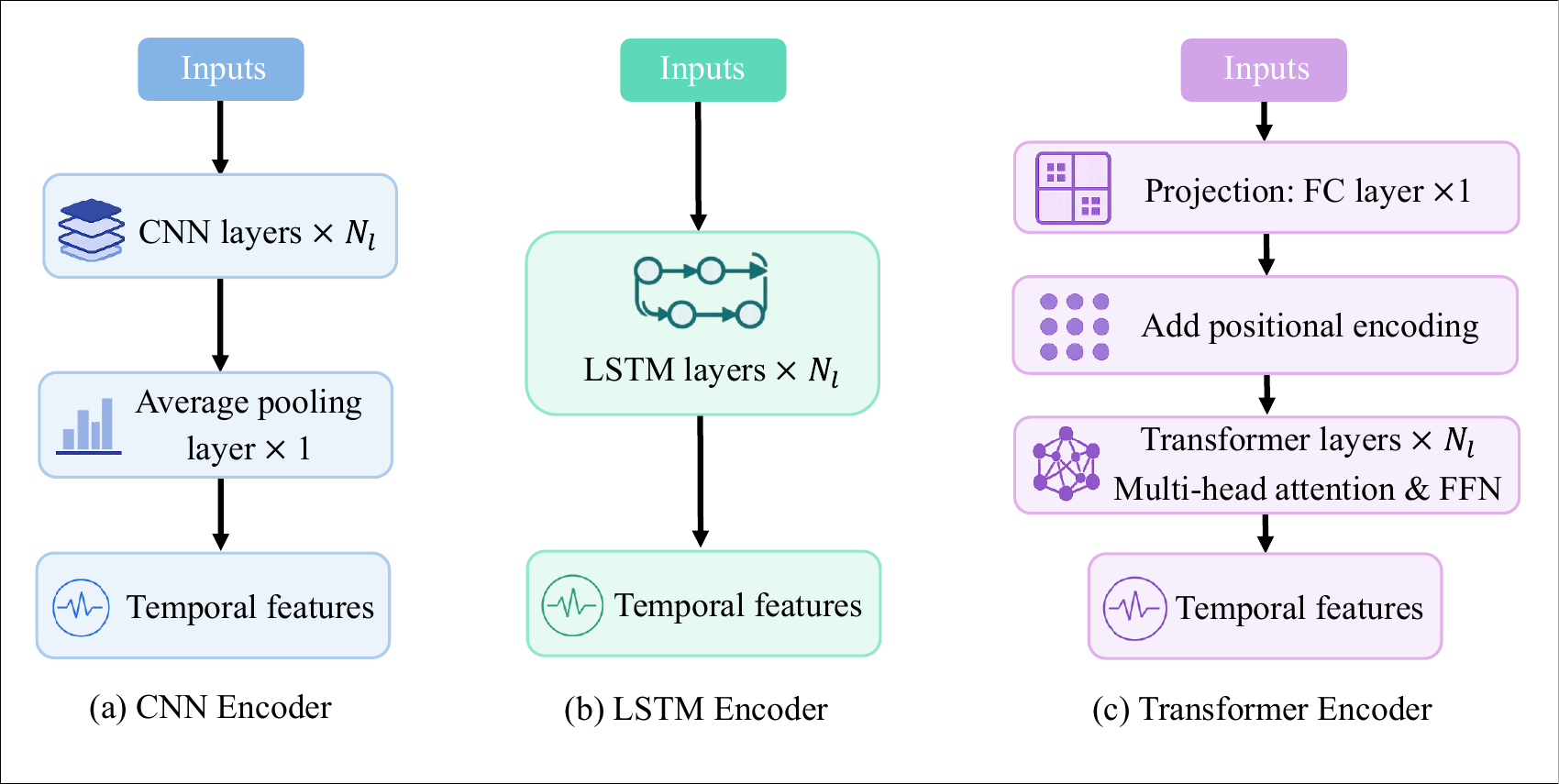}
		\caption{Three typical temporal encoders} 
		\label{fig:2}
	\end{center}
\end{figure}

\textbf{Encoder Design:} To demonstrate the adaptability of the proposed framework across different temporal modelling paradigms, three widely used encoder architectures (CNN, LSTM, and Transformer) were employed to extract sequential features from the input time series, structured as in Figure \ref{fig:2}. It is worth emphasizing that the temporal encoder is not the focus of innovation in this work. Instead, these encoder variants are incorporated to validate the robustness and generality of the proposed task decomposition and adaptive learning mechanisms under diverse feature extraction backbones.

\textit{CNN Encoder:} The CNN-based encoder consists of $N_l$ stacked convolutional layers followed by an average pooling layer, which extracts compact temporal features from the input sequence.

\textit{LSTM Encoder:} The LSTM-based encoder is composed of $N_l$ stacked LSTM layers, which encode temporal dependencies by processing the input sequence into hidden representations.

\textit{Transformer Encoder:} The Transformer-based encoder first projects the input into a unified feature space and adds positional encodings. The resulting representations are then processed by $N_l$ Transformer encoder layers with multi-head self-attention and feed-forward networks.

\textbf{Baselines:} To enable fair and transparent comparisons, three reference baseline models are designed, as illustrated in Figure \ref{fig:A.5-1}. Specifically, Baseline3 adopts a structure closely aligned with CaReTS, but replaces the proposed trend--deviation fusion formulation with a single fully connected layer. Baseline2 further simplifies Baseline3 by removing the residual connection, such that the network directly outputs $\hat{\mathbf{y}}$ instead of modelling deviations relative to the latest observation $x_n$. Finally, Baseline1 represents a conventional encoder--decoder design, where the encoder (CNN, LSTM, or Transformer layers) is followed by $N_b$ fully connected layers that directly map the input sequence to multi-step predictions. All baseline models are formulated as single-task regression models and are constructed with progressively simplified structures. This design provides a set of comparable reference models for evaluating the proposed CaReTS framework, enabling a systematic comparison between the multi-task decomposition approach and conventional single-task forecasting models under similar or simplified architectural settings.
\textbf{\begin{figure}[t]
	\begin{center}
		\includegraphics[width=16cm, trim=5 5 5 5, clip]{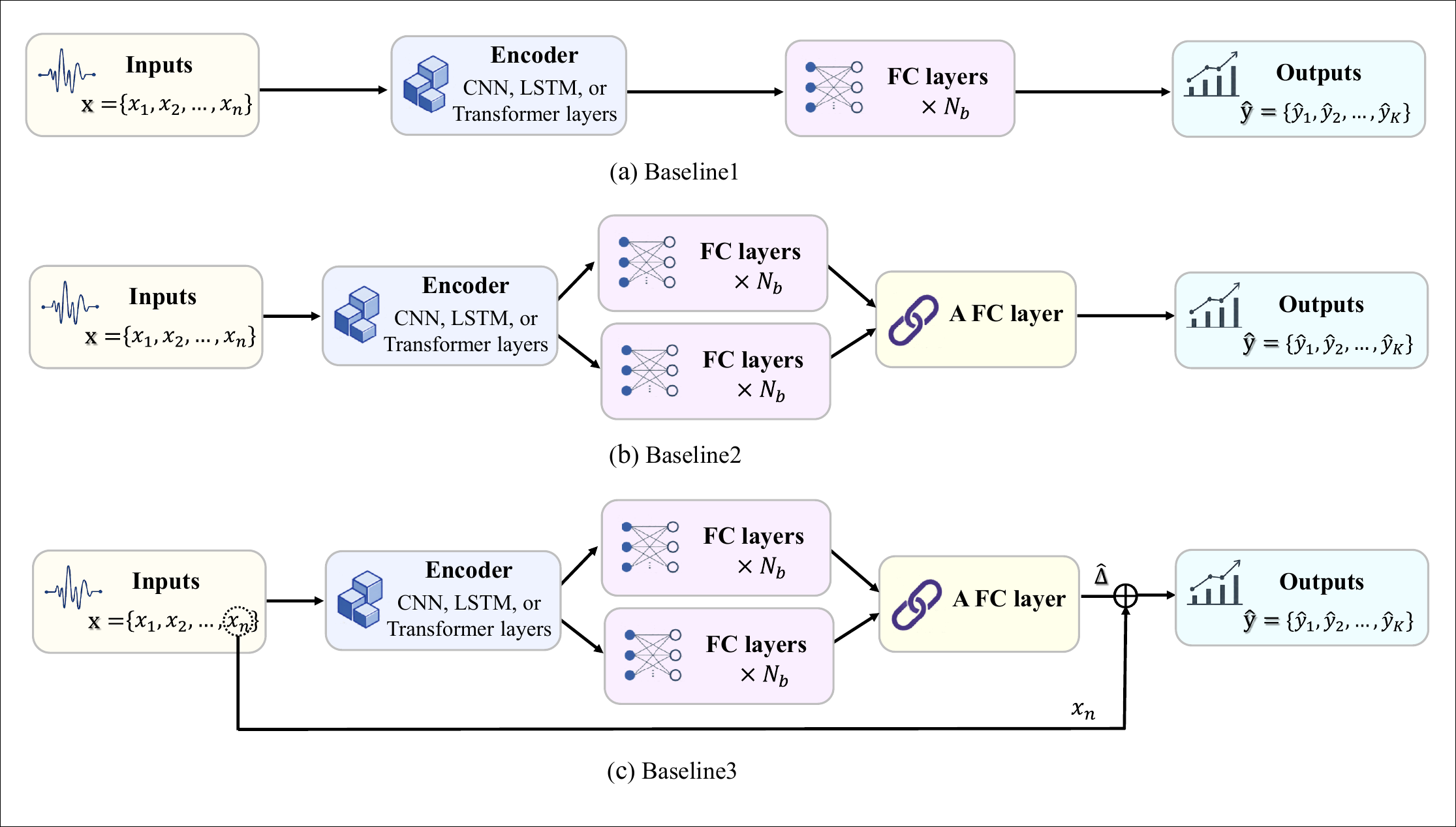}    
           \captionsetup{skip=8 pt}
		\caption{Structures of three new baselines} 
		\label{fig:A.5-1}
	\end{center}
\end{figure}}

\textbf{Configuration:} All experiments were implemented in Python and executed on Google Colab equipped with a single NVIDIA T4 GPU. The proposed CaReTS framework employed two fully connected layers with 64 hidden units in both the trend classification branch and the deviation estimation branch. Model training was conducted for up to 600 epochs, with early stopping applied if no performance improvement was observed over 50 consecutive epochs. The Adam optimizer was used with a learning rate of 0.001, and the batch size was set to 64. To ensure reproducibility, the random seed was fixed at 2025. All input data were normalized using Min-Max scaling, and the ReLU activation function was adopted throughout the network. For the encoder design, $N_l=2$ layers with 64 hidden units were adopted in three encoder variants. Specifically, a kernel size of 3 with padding of 1 was used in the CNN encoder, while all Transformer encoders were configured with 4 attention heads. For the Baseline1–3 algorithms, we also set $N_b=N_l=2$ with fully connected layers of 64 units each.

\textbf{Evaluation Protocol:} To ensure robust and reliable evaluation, a 10-run training–validation procedure was performed on the training set, while the test set was strictly reserved for final evaluation and not used during model development. In each run, the training data were divided into 10 folds, where 9 folds were used for training and the remaining fold was used for validation. A new model was initialized and trained in each run, resulting in 10 independently trained models. Each model was then evaluated on the held-out test set, and the final results were reported as the mean and standard deviation across the 10 models. This protocol ensures that the test set remains completely isolated from the training and validation processes, thereby avoiding any data leakage in the evaluation.

\subsection{Architecture Evaluation}\label{sec:4.2}

\begin{figure*}[htbp]
    \centering
    \setlength{\abovecaptionskip}{-3 pt} 
    \begin{minipage}{0.48\textwidth}
        \centering
        \includegraphics[width=\textwidth]{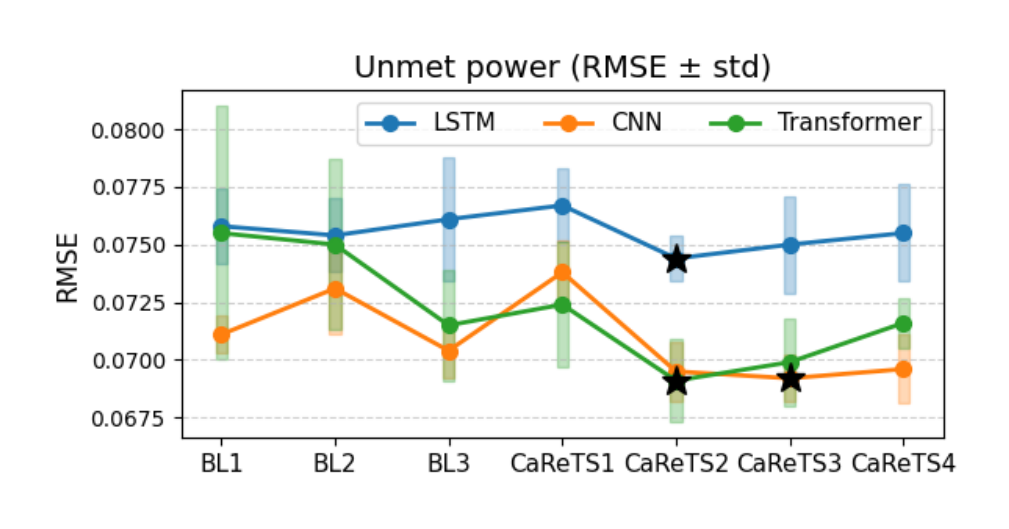}
        \caption{RMSE on power across approaches}
        \label{fig:3}
    \end{minipage}\hfill
    \begin{minipage}{0.48\textwidth}
        \centering
        \includegraphics[width=\textwidth]{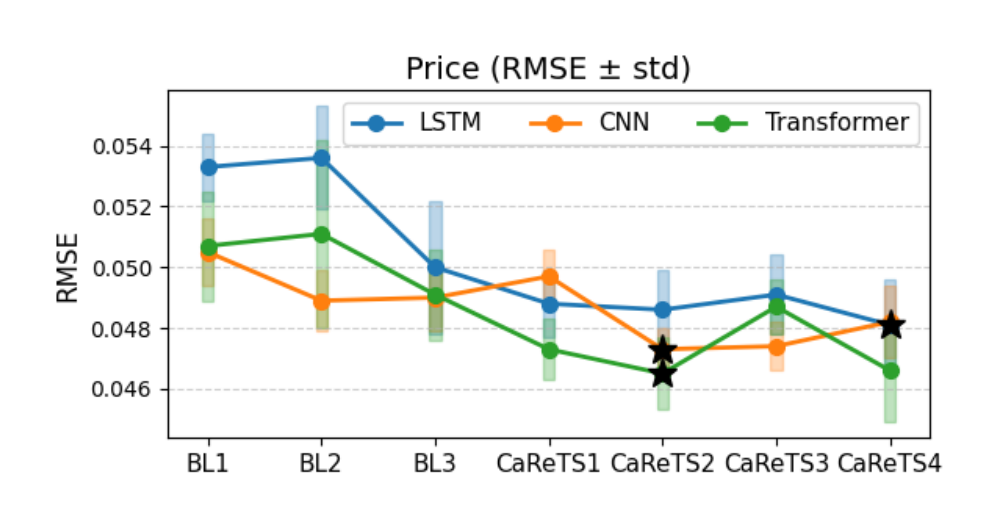}
        \caption{RMSE on price across approaches}
        \label{fig:4}
    \end{minipage}
\end{figure*}

We first evaluate CaReTS 1–4 with three encoders (i.e., CNN, LSTM, and Transformer) on the electricity price and unmet power forecasting tasks. Following the original setup \cite{yao2025self}, both tasks adopted a 15-to-6 forecasting scheme, where the inputs consisted of temporal features (month, weekday, and hour) at the current time step along with the previous 12 observations of the target variable, and the outputs corresponded to predictions over the next 6 time steps. Additional analyses under other input–output configurations are provided in Section \ref{sec:4.7}. Table \ref{tab:2} reports the average RMSE for 6-step forecasting (mean ± standard deviation over 10 runs), while Figures \ref{fig:3} and \ref{fig:4} illustrate the corresponding RMSE results on the test set. Here, BL1, BL2, and BL3 denote Baseline1, Baseline2, and Baseline3, respectively, and ‘$\bigstar$’ indicates the best-performing model (i.e., the lowest RMSE) under each encoder.

\begin{table}[!b]
\centering
\caption{Average RMSE (mean ± std) for 6-step forecasting across approaches (test set)}
\label{tab:2}
\resizebox{\textwidth}{!}
{%
\begin{tabular}{lcccccc}
\hline
\textbf{Approach} &
\multicolumn{3}{c}{\textbf{Unmet power}} &
\multicolumn{3}{c}{\textbf{Electricity price}} \\
\cmidrule(lr){2-4} \cmidrule(lr){5-7}
 & Train & Validation & Test & Train & Validation & Test \\
\hline
\multicolumn{7}{c}{\textbf{LSTM}} \\
Baseline1 & 0.0460 $\pm$ 0.0040 & 0.0666 $\pm$ 0.0030 & 0.0758 $\pm$ 0.0016 & 0.0198 $\pm$ 0.0017 & 0.0378 $\pm$ 0.0027 & 0.0533 $\pm$ 0.0011 \\
Baseline2 & 0.0454 $\pm$ 0.0046 & 0.0691 $\pm$ 0.0016 & 0.0754 $\pm$ 0.0016 & 0.0215 $\pm$ 0.0019 & 0.0423 $\pm$ 0.0018 & 0.0536 $\pm$ 0.0017 \\
Baseline3 & 0.0453 $\pm$ 0.0049 & 0.0682 $\pm$ 0.0021 & 0.0761 $\pm$ 0.0027 & 0.0218 $\pm$ 0.0015 & 0.0393 $\pm$ 0.0013 & 0.0500 $\pm$ 0.0022 \\
CaReTS1   & 0.0550 $\pm$ 0.0033 & 0.0723 $\pm$ 0.0021 & 0.0767 $\pm$ 0.0016 & 0.0265 $\pm$ 0.0020 & 0.0427 $\pm$ 0.0017 & 0.0488 $\pm$ 0.0011 \\
CaReTS2   & 0.0512 $\pm$ 0.0023 & 0.0684 $\pm$ 0.0025 & \textbf{0.0744 $\pm$ 0.0010} & 0.0262 $\pm$ 0.0019 & 0.0400 $\pm$ 0.0025 & 0.0486 $\pm$ 0.0013 \\
CaReTS3   & 0.0474 $\pm$ 0.0036 & 0.0691 $\pm$ 0.0030 & 0.0750 $\pm$ 0.0021 & 0.0240 $\pm$ 0.0018 & 0.0408 $\pm$ 0.0020 & 0.0491 $\pm$ 0.0013 \\
CaReTS4   & 0.0519 $\pm$ 0.0032 & 0.0714 $\pm$ 0.0019 & 0.0755 $\pm$ 0.0021 & 0.0314 $\pm$ 0.0014 & 0.0435 $\pm$ 0.0024 & \textbf{0.0481 $\pm$ 0.0015} \\
\hline
\multicolumn{7}{c}{\textbf{CNN}} \\
Baseline1 & 0.0619 $\pm$ 0.0024 & 0.0679 $\pm$ 0.0022 & 0.0711 $\pm$ 0.0008 & 0.0410 $\pm$ 0.0015 & 0.0463 $\pm$ 0.0018 & 0.0505 $\pm$ 0.0011 \\
Baseline2 & 0.0550 $\pm$ 0.0026 & 0.0661 $\pm$ 0.0021 & 0.0731 $\pm$ 0.0020 & 0.0317 $\pm$ 0.0015 & 0.0408 $\pm$ 0.0017 & 0.0489 $\pm$ 0.0010 \\
Baseline3 & 0.0576 $\pm$ 0.0020 & 0.0655 $\pm$ 0.0020 & 0.0704 $\pm$ 0.0012 & 0.0310 $\pm$ 0.0019 & 0.0413 $\pm$ 0.0018 & 0.0490 $\pm$ 0.0011 \\
CaReTS1   & 0.0696 $\pm$ 0.0012 & 0.0739 $\pm$ 0.0029 & 0.0738 $\pm$ 0.0014 & 0.0443 $\pm$ 0.0013 & 0.0499 $\pm$ 0.0013 & 0.0497 $\pm$ 0.0009 \\
CaReTS2   & 0.0658 $\pm$ 0.0013 & 0.0694 $\pm$ 0.0025 & 0.0695 $\pm$ 0.0013 & 0.0427 $\pm$ 0.0010 & 0.0470 $\pm$ 0.0017 & \textbf{0.0473 $\pm$ 0.0007} \\
CaReTS3   & 0.0609 $\pm$ 0.0020 & 0.0665 $\pm$ 0.0022 & \textbf{0.0692 $\pm$ 0.0010} & 0.0377 $\pm$ 0.0013 & 0.0443 $\pm$ 0.0015 & 0.0474 $\pm$ 0.0008 \\
CaReTS4   & 0.0626 $\pm$ 0.0018 & 0.0678 $\pm$ 0.0022 & 0.0696 $\pm$ 0.0015 & 0.0428 $\pm$ 0.0014 & 0.0475 $\pm$ 0.0018 & 0.0482 $\pm$ 0.0012 \\
\hline
\multicolumn{7}{c}{\textbf{Transformer}} \\
Baseline1 & 0.0561 $\pm$ 0.0084 & 0.0683 $\pm$ 0.0066 & 0.0755 $\pm$ 0.0055 & 0.0322 $\pm$ 0.0025 & 0.0412 $\pm$ 0.0028 & 0.0507 $\pm$ 0.0018 \\
Baseline2 & 0.0530 $\pm$ 0.0056 & 0.0683 $\pm$ 0.0036 & 0.0750 $\pm$ 0.0037 & 0.0359 $\pm$ 0.0037 & 0.0436 $\pm$ 0.0031 & 0.0511 $\pm$ 0.0031 \\
Baseline3 & 0.0542 $\pm$ 0.0044 & 0.0667 $\pm$ 0.0030 & 0.0715 $\pm$ 0.0024 & 0.0353 $\pm$ 0.0037 & 0.0443 $\pm$ 0.0034 & 0.0491 $\pm$ 0.0015 \\
CaReTS1   & 0.0583 $\pm$ 0.0022 & 0.0702 $\pm$ 0.0028 & 0.0724 $\pm$ 0.0027 & 0.0341 $\pm$ 0.0016 & 0.0444 $\pm$ 0.0026 & 0.0473 $\pm$ 0.0010 \\
CaReTS2   & 0.0588 $\pm$ 0.0016 & 0.0686 $\pm$ 0.0016 & \textbf{0.0691 $\pm$ 0.0018} & 0.0333 $\pm$ 0.0020 & 0.0445 $\pm$ 0.0027 & \textbf{0.0465 $\pm$ 0.0012} \\
CaReTS3   & 0.0536 $\pm$ 0.0026 & 0.0665 $\pm$ 0.0022 & 0.0699 $\pm$ 0.0019 & 0.0327 $\pm$ 0.0017 & 0.0428 $\pm$ 0.0028 & 0.0487 $\pm$ 0.0009 \\
CaReTS4   & 0.0588 $\pm$ 0.0031 & 0.0696 $\pm$ 0.0024 & 0.0716 $\pm$ 0.0011 & 0.0375 $\pm$ 0.0027 & 0.0453 $\pm$ 0.0022 & 0.0466 $\pm$ 0.0017 \\
\hline
\end{tabular}
} 
\end{table}

The results show that the proposed CaReTS 2–4 models outperformed all baselines across both forecasting tasks on the test set (regardless of the encoder employed), indicating the effectiveness of the proposed multi-task decomposition framework. Among these variants, CaReTS2 achieved the best overall performance, yielding the lowest RMSE in four out of six cases and the second-best results in the remaining cases. In particular, CaReTS2 combined with the Transformer encoder achieved the lowest RMSE of 0.0691 $\pm$ 0.0018 for unmet power and 0.0465 $\pm$ 0.0012 for electricity price, highlighting the strong synergy between the proposed framework and attention-based temporal modelling. Interestingly, although the proposed models did not always achieve the lowest errors on the training and validation sets, they consistently yielded better performance on the test set. This behavior may reflect improved generalization, suggesting that the proposed multi-task decomposition helps reduce overfitting and improves robustness under unseen data. However, CaReTS1 did not demonstrate a clear advantage over the baseline models. This may be associated with the simplified design of its deviation branch, which limits its ability to effectively capture the discrepancy between predicted trends and actual observations. In contrast, the more complete formulations in CaReTS2–4 better exploit the decomposition structure, leading to more stable performance.

A comparison with the baseline models further provides insights into the contribution of different components. The baseline variants were constructed by progressively simplifying the model from Baseline3 to Baseline1. Although performance did not degrade strictly monotonically, the simplified variants generally exhibited inferior results compared to CaReTs models. These observations suggest that both the decomposition strategy and the residual modelling contribute to improving forecasting performance. Moreover, the performance advantage of CaReTS is observed across all three encoder types, indicating that the proposed framework is compatible with different temporal feature extraction mechanisms, including convolutional, recurrent, and attention-based models. In the context of energy forecasting, such improvements may reflect a better ability to capture short-term fluctuations in price and load dynamics, which are critical for real-time operational decisions.

\subsection{Trend Prediction and Analysis}\label{sec:4.3}

To evaluate the effectiveness of the proposed framework in capturing directional dynamics, we analyze the trend prediction accuracy of the classification branch. Table \ref{tab:3} reports the average accuracy for multi-step forecasting using CaReTS1–4 with different encoder architectures. For CaReTS3 and CaReTS4, the predicted trend is determined by selecting the direction with the higher probability ($P_{up}$ or $P_{down}$), which is used to compute the classification accuracy. All CaReTS variants achieved accuracy above 90\% across both tasks, indicating that the proposed framework is highly effective in capturing the directional dynamics of time series. 
Among the encoder architectures, the Transformer consistently outperformed CNN and LSTM, indicating its stronger capability in capturing temporal dependencies within the proposed multi-task decomposition framework. Consistently, the CaReTS2–Transformer combination achieved the highest accuracy, aligning with its superior RMSE performance reported in Table \ref{tab:2}. This result suggests that explicit directional supervision is compatible with numerical forecasting and can provide an additional diagnostic output for inspecting predicted short-term movements.

\begin{table}[t]
\centering
\caption{Average trend accuracy for multi-step forecasting across approaches (test set)}
\label{tab:3}
\setlength{\tabcolsep}{6pt} 
\renewcommand{\arraystretch}{1.0} 
\scriptsize 
\begin{tabular}{lcccc}
\hline
\textbf{Encoder} & \textbf{CaReTS1} & \textbf{CaReTS2} & \textbf{CaReTS3} & \textbf{CaReTS4} \\
\hline
\multicolumn{5}{c}{\textbf{Unmet power}} \\
LSTM        & 0.9111 $\pm$ 0.0020 & 0.9096 $\pm$ 0.0019 & 0.9086 $\pm$ 0.0030 & 0.9068 $\pm$ 0.0041 \\
CNN         & 0.9125 $\pm$ 0.0032 & 0.9127 $\pm$ 0.0033 & 0.9125 $\pm$ 0.0032 & 0.9140 $\pm$ 0.0020 \\
Transformer & 0.9191 $\pm$ 0.0032 & \textbf{0.9192 $\pm$ 0.0022} & 0.9168 $\pm$ 0.0029 & 0.9166 $\pm$ 0.0025 \\

\hline
\multicolumn{5}{c}{\textbf{Electricity price}} \\
LSTM        & 0.9073 $\pm$ 0.0027 & 0.9071 $\pm$ 0.0030 & 0.9066 $\pm$ 0.0032 & 0.9056 $\pm$ 0.0038 \\
CNN         & 0.9032 $\pm$ 0.0015 & 0.9036 $\pm$ 0.0030 & 0.9024 $\pm$ 0.0041 & 0.9016 $\pm$ 0.0043 \\
Transformer & 0.9142 $\pm$ 0.0029 & \textbf{0.9146 $\pm$ 0.0019} & 0.9135 $\pm$ 0.0021 & 0.9136 $\pm$ 0.0051 \\
\hline
\end{tabular}
\end{table}

\begin{figure*}[htbp]
    \setlength{\abovecaptionskip}{-7pt} 
    \centering
    \begin{minipage}{0.48\textwidth}
        \centering
        \includegraphics[width=\textwidth]{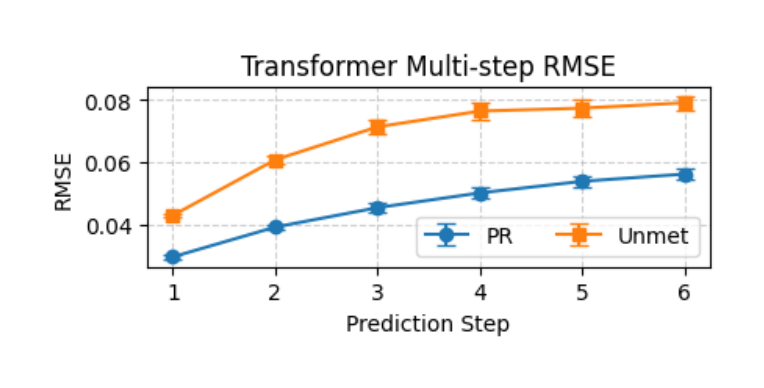}
        \caption{RMSE across forecasting steps using CaReTS2-Transformer}
        \label{fig:5}
    \end{minipage}\hfill
    \begin{minipage}{0.48\textwidth}
        \centering
        \includegraphics[width=\textwidth]{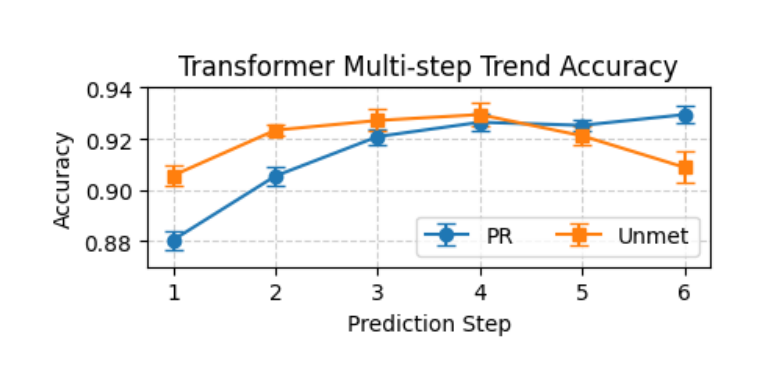}
        \caption{Trend accuracy across forecasting steps using CaReTS2-Transformer}
        \label{fig:6}
    \end{minipage}
\end{figure*}

Figures \ref{fig:5} and \ref{fig:6}, obtained using this best-performing CaReTS2–Transformer model on test set, further illustrate the evolution of classification accuracy and RMSE across six forecasting steps. As expected, RMSE gradually increased with longer forecast horizons due to error accumulation, consistent with prior studies \citep{yao2025self,yunpeng2017multi,venkatraman2015improving}. Interestingly, trend classification accuracy did not exhibit a declining pattern, indicating the robustness of the proposed framework in maintaining reliable trend detection even over longer forecasting horizons. This stability can be attributed to the complementary design of the framework: the classification branch captures macro-level trend directions to provide stable directional guidance, while the regression branch refines micro-level numerical predictions. 
From an energy perspective, such stable directional predictions are particularly valuable, as they provide reliable signals for short-term decision-making even when numerical uncertainty increases.

\subsection{Multi-Task Learning and Mechanism Ablation}\label{sec:4.4}

Apart from the structural comparisons above, we further investigate the multi-task learning mechanism of the proposed framework through some ablation analysis. We adopted the Transformer encoder as a representative case, as it consistently achieved the best performance in the preceding experiments. In addition to the proposed uncertainty-aware weighting scheme, we also introduced a fixed-weight multi-task setting and a single-task setting for comparison. In the single-task setting, the same backbone network was used, but only the output prediction loss $L_{op}$ was optimized, while the classification $L_{ca}$ and deviation $L_{de}$ losses were removed. In the fixed-weight multi-task setting, all task losses were jointly optimized with equal weights (i.e., $L_{ca}:L_{de}:L_{op}=1:1:1$ for CaReTS1–3 and $L_{ca}:L_{op}=1:1$ for CaReTS4), allowing us to isolate the effect of adaptive task balancing. Table \ref{tab:4} reports the comparison among the proposed adaptive multi-task learning, fixed-weight multi-task learning, and single-task learning. The runtime, measured in seconds (s), is reported as the average per run across the 10-run training–validation procedure.

The adaptive multi-task setting achieved better and more stable performance than both the fixed-weight multi-task setting and the single-task setting across both forecasting tasks, suggesting that joint optimization could provide complementary learning rather than task interference. In contrast, the fixed-weight multi-task setting exhibited less consistent behavior. Generally, the fixed-weight setting tended to result in higher RMSE and lower trend accuracy compared with the adaptive scheme, and in some cases performed even worse than the single-task setting. This may be due to differences in loss scales, convergence rates, and noise characteristics across tasks, which can lead to suboptimal task balancing under fixed weighting. As a result, certain tasks may sometimes dominate the training process, leading to less stable performance and increased training time. By contrast, the uncertainty-aware weighting scheme adaptively balanced task contributions during training, leading to more stable performance across both numerical forecasting and directional prediction tasks.
Regarding trend prediction, the adaptive multi-task setting consistently achieved accuracy above 91\%, outperforming both fixed-weight and single-task settings. This confirms that explicitly modeling trend classification provides reliable directional guidance, which complements numerical forecasting. 

\begin{table}[t]
\centering
\caption{Comparison of adaptive multi-task, fixed-weight multi-task, and single-task learning (Transformer encoder)}
\label{tab:4}
\setlength{\tabcolsep}{1.5pt} 
\renewcommand{\arraystretch}{1.1}
\scriptsize

\resizebox{\textwidth}{!}{%
\begin{tabular}{lccccccccc}
\hline
\textbf{Model} &
\multicolumn{3}{c}{\textbf{Adaptive multi-task}} &
\multicolumn{3}{c}{\textbf{Fixed-weight multi-task}} &
\multicolumn{3}{c}{\textbf{Single-task}} \\
\cmidrule(lr){2-4} \cmidrule(lr){5-7} \cmidrule(lr){8-10}

 & RMSE & Trend Acc. & Time & RMSE & Trend Acc. & Time & RMSE & Acc. & Time \\
\hline

\multicolumn{10}{c}{\textbf{Unmet power}} \\

CaReTS1 & 0.0724$\pm$0.0027 & 0.9191$\pm$0.0032 & 253 &
           0.0766$\pm$0.0052 & 0.9071$\pm$0.0071 & 281 &
           0.0758$\pm$0.0036 & 0.8874$\pm$0.0046 & 216 \\

CaReTS2 & \textbf{0.0691}$\pm$\textbf{0.0018} & \textbf{0.9192}$\pm$\textbf{0.0022} & 256 &
           \textbf{0.0691}$\pm$\textbf{0.0031} & 0.9163$\pm$0.0047 & 248 &
           0.0704$\pm$0.0029 & 0.9060$\pm$0.0023 & 261 \\

CaReTS3 & 0.0699$\pm$0.0019 & 0.9168$\pm$0.0029 & 296 &
           0.0728$\pm$0.0048 & 0.9112$\pm$0.0062 & 334 &
           0.0721$\pm$0.0017 & 0.8965$\pm$0.0036 & 236 \\

CaReTS4 & 0.0716$\pm$0.0011 & 0.9166$\pm$0.0025 & 306 &
           0.0724$\pm$0.0039 & 0.9044$\pm$0.0063 & 318 &
           0.0716$\pm$0.0026 & 0.9053$\pm$0.0058 & 313 \\

\hline
\multicolumn{10}{c}{\textbf{Electricity price}} \\

CaReTS1 & 0.0473$\pm$0.0010 & 0.9142$\pm$0.0029 & 357 &
           0.0492$\pm$0.0038 & 0.9088$\pm$0.0061 & 398 &
           0.0539$\pm$0.0023 & 0.8663$\pm$0.0028 & 333 \\

CaReTS2 & \textbf{0.0465}$\pm$\textbf{0.0012} & \textbf{0.9146}$\pm$\textbf{0.0019} & 388 &
           0.0470$\pm$0.0041 & 0.9117$\pm$0.0053 & 374 &
           0.0470$\pm$0.0020 & 0.8939$\pm$0.0033 & 379 \\

CaReTS3 & 0.0487$\pm$0.0009 & 0.9135$\pm$0.0021 & 401 &
           0.0503$\pm$0.0038 & 0.8851$\pm$0.0067 & 451 &
           0.0474$\pm$0.0012 & 0.8860$\pm$0.0014 & 386 \\

CaReTS4 & 0.0466$\pm$0.0017 & 0.9136$\pm$0.0051 & 321 &
           0.0470$\pm$0.0033 & 0.9141$\pm$0.0071 & 342 &
           0.0472$\pm$0.0018 & 0.8889$\pm$0.0042 & 318 \\

\hline
\end{tabular}
}
\end{table}

\subsection{Comparison with SOTA Algorithms}\label{sec:4.5}

\begin{figure*}[t]
    \setlength{\abovecaptionskip}{-7pt} 
    \centering
    \begin{minipage}{0.48\textwidth}
        \centering
        \includegraphics[width=\textwidth, trim=5 5 5 5, clip]{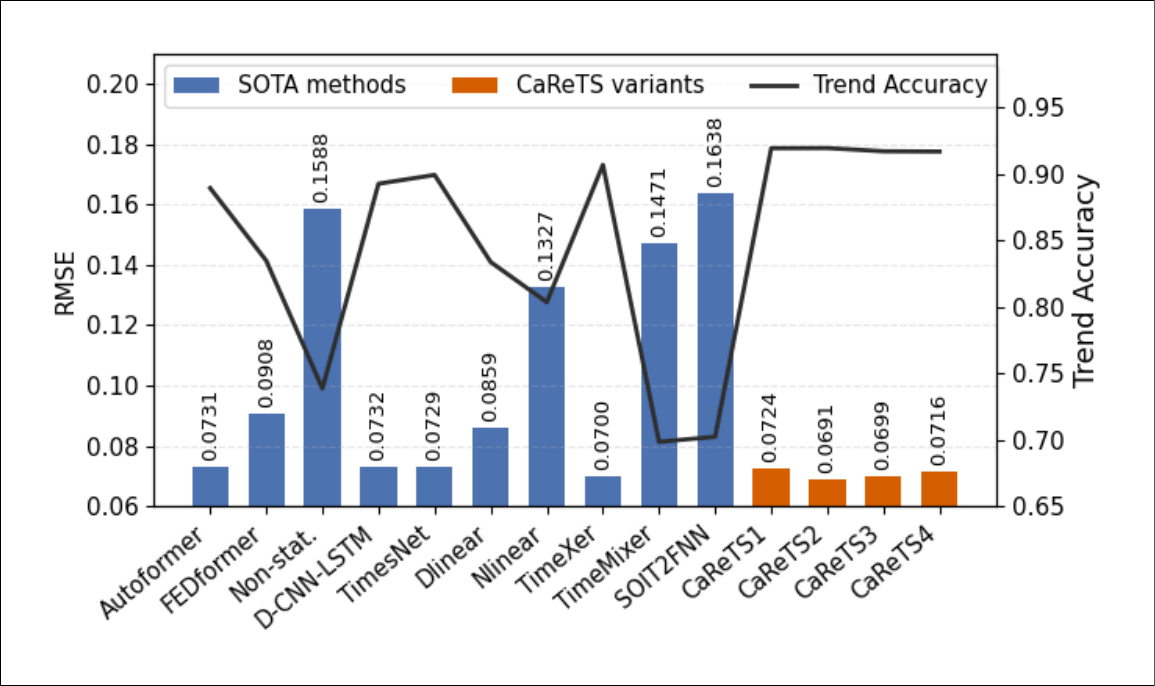}
        \caption{Comparison of RMSE and Trend Accuracy on Unmet Power Forecasting}
        \label{fig:4.5-1}
    \end{minipage}\hfill
    \begin{minipage}{0.48\textwidth}
        \centering
        \includegraphics[width=\textwidth, trim=5 5 5 5, clip]{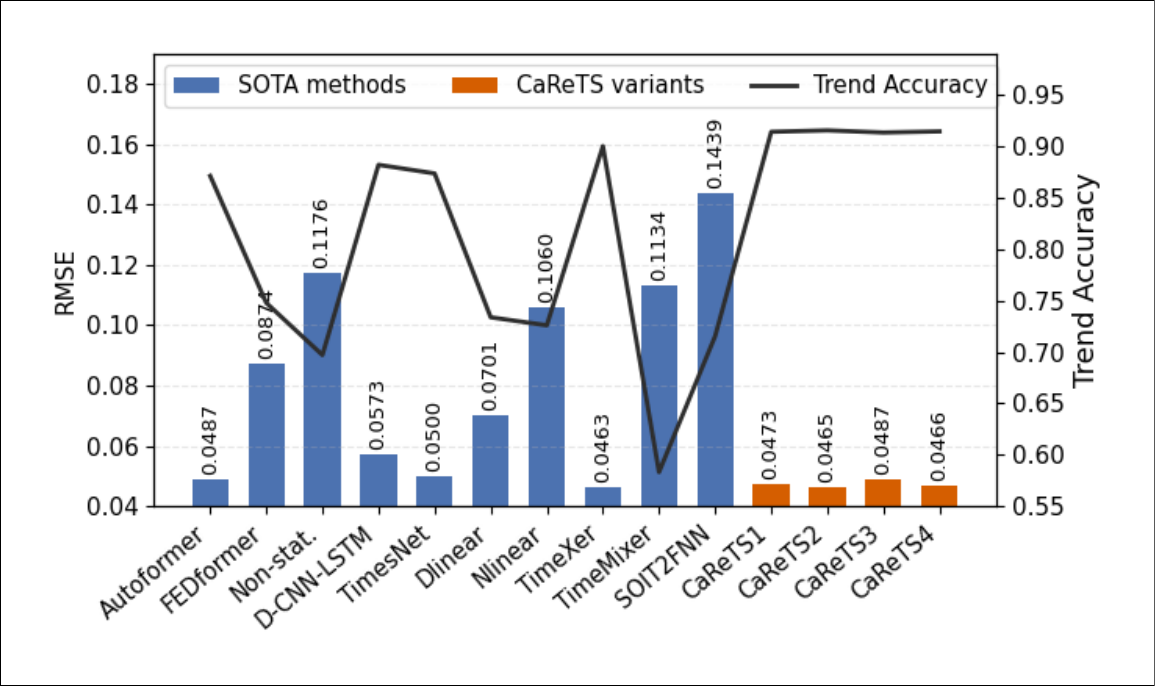}
        \caption{Comparison of RMSE and Trend Accuracy on Electricity Price Forecasting}
        \label{fig:4.5-2}
    \end{minipage}
\end{figure*}

Table \ref{tab:5} summarizes the results of ten representative state-of-the-art (SOTA) algorithms, which are compared against the proposed adaptive multi-task CaReTS variants (Table \ref{tab:4}). The corresponding visual comparisons of RMSE and trend accuracy for unmet power and electricity price forecasting are illustrated in Figures \ref{fig:4.5-1} and \ref{fig:4.5-2}, respectively. The results are reported under the 15-input–6-output setting, while additional results under alternative input–output configurations are provided in Section \ref{sec:4.7}. The comparison clearly shows that CaReTS achieved state-of-the-art performance, particularly in reducing RMSE while maintaining consistently strong trend prediction accuracy. For unmet power forecasting, CaReTS2 and CaReTS3 yielded the lowest RMSE values (0.0691 and 0.0699, respectively) with trend accuracy above 0.916, outperforming the best SOTA model TimeXer. All CaReTS variants remained competitive, with even the simpler configurations (CaReTS1 and CaReTS4) surpassing most SOTA models. For electricity price forecasting, the CaReTS family again showed strong and consistent performance. While TimeXer achieved comparable RMSE, CaReTS2 provided a better overall balance by combining competitive prediction accuracy (RMSE = 0.0465) with the highest trend accuracy (0.9146). In contrast, TimeXer exhibited lower trend accuracy (0.9013) and significantly higher computational cost. A key advantage of the proposed CaReTS framework lies in its consistently high trend prediction accuracy across all variants. This can be attributed to the multi-task design, where trend classification is explicitly modeled, enabling the model to capture directional dynamics more effectively while improving interpretability. This is particularly beneficial in energy applications, where consistent directional signals can improve decision robustness even when numerical differences are small.

\begin{table}[t]
\centering
\caption{Test results of SOTA algorithms on unmet power and electricity price forecasting}
\label{tab:5}
\resizebox{\textwidth}{!}{%
\begin{tabular}{lccccccc}
\hline
\textbf{Approach} &
\multicolumn{3}{c}{\textbf{Unmet power}} &
\multicolumn{3}{c}{\textbf{Electricity price}} \\
\cmidrule(lr){2-4} \cmidrule(lr){6-8} 
 & RMSE & Trend Acc. & Time (s) & & RMSE & Trend Acc. & Time (s) \\
\hline
Autoformer \citep{wu2021autoformer}& 0.0731 $\pm$ 0.0009 & 0.8891 $\pm$ 0.0036 & 510.05  & & 0.0487 $\pm$ 0.0021 & 0.8713 $\pm$ 0.0073 & 467.97 \\
FEDformer \citep{zhou2022fedformer}  & 0.0908 $\pm$ 0.0005 & 0.8345 $\pm$ 0.0023 & 222.80 &  & 0.0874 $\pm$ 0.0011 & 0.7477 $\pm$ 0.0086 & 239.34 \\
Non-stationary Tr \citep{liu2022non}& 0.1588 $\pm$ 0.0025 & 0.7384 $\pm$ 0.0076 & 541.35  & & 0.1176 $\pm$ 0.0036 & 0.6970 $\pm$ 0.0172 & 422.41 \\
D-CNN-LSTM\citep{yao2022integrated} & 0.0732 $\pm$ 0.0009 & 0.8924 $\pm$ 0.0034 & 103.28  & & 0.0573 $\pm$ 0.0012 & 0.8821 $\pm$ 0.0067 & 112.02 \\
TimesNet \citep{wu2023timesnet}& 0.0729 $\pm$ 0.0012 & 0.8990 $\pm$ 0.0028 & 273.15  & & 0.0500 $\pm$ 0.0015 & 0.8737 $\pm$ 0.0074 & 314.40  \\
Dlinear \citep{zeng2023transformers}   & 0.0859 $\pm$ 0.0004 & 0.8335 $\pm$ 0.0028 & 68.75  &  & 0.0701 $\pm$ 0.0005 & 0.7337 $\pm$ 0.0085 & 70.07 \\
Nlinear \citep{zeng2023transformers}    & 0.1327 $\pm$ 0.0002 & 0.8033 $\pm$ 0.0017 & 48.44  &  & 0.1060 $\pm$ 0.0004 & 0.7259 $\pm$ 0.0036 & 50.33 \\
TimeXer \citep{wang2024timexer}& 0.0700 $\pm$ 0.0022 & 0.9066 $\pm$ 0.0022 & 448.62   & & 0.0463 $\pm$ 0.0013 & 0.9013 $\pm$ 0.0054 & 573.75  \\
TimeMixer \citep{wang2024timemixer}& 0.1471 $\pm$  0.0008 & 0.6983 $\pm$  0.0048 & 76.25   & & 0.1134 $\pm$  0.0010 & 0.5831 $\pm$ 0.0100 & 84.60  \\
SOIT2FNN-MO \citep{yao2025self}& 0.1638 $\pm$ 0.0012 & 0.7021 $\pm$ 0.0020 & 863.05  & & 0.1439 $\pm$ 0.0018 & 0.7153 $\pm$ 0.0042 & 926.81 \\
\hline
\end{tabular}
} 
\end{table}

From an efficiency perspective, CaReTS ran within a moderate cost ($\approx$200–400s), which was much faster than heavier architectures such as Autoformer ($>$460s) or SOIT2FNN-MO ($>$860s). Although slower than lightweight baselines (e.g., Nlinear/Dlinear $<$70s and TimeMixer $<85$s), CaReTS achieved a favorable trade-off, where the additional computation was modest compared to the substantial accuracy gains. Furthermore, the proposed framework showed strong potential for reliable multi-step forecasting. As illustrated in Figure \ref{fig:6}, trend prediction accuracy remained stable across increasing forecasting horizons, without the typical degradation. This may provide an insightful indicator that supports the applicability of CaReTS algorithms to situations requiring extended periods of forecasting. 

\subsection{Alternative Input–Output Settings} \label{sec:4.7}

We extend the evaluation to additional input–output settings to examine the robustness of the proposed framework under varying forecasting horizons. Tables \ref{tab:A.6a} and \ref{tab:A.6b} present the results on the unmet power and electricity price datasets under the 15-input–4-output and 15-input–8-output settings, respectively. For each configuration, the lowest three RMSE values and the highest three trend accuracies among all methods are highlighted in bold. Overall, the proposed CaReTS variants maintained consistently strong performance across different forecasting horizons. In particular, they demonstrated clear advantages in trend prediction accuracy, indicating the effectiveness of the proposed multi-task decomposition in capturing directional dynamics. Across both datasets and settings, CaReTS2 consistently ranked among the top-performing models in terms of both RMSE and trend accuracy, suggesting a favorable balance between numerical prediction and directional consistency. Notably, the performance of CaReTS remained stable when the prediction horizon was extended to 8 steps or reduced to 4 steps, indicating strong generalization capability under more challenging forecasting scenarios.

\begin{table*}[t]
\centering
\caption{Comparison of SOTA algorithms for 15-4 multi-step forecasting}
\label{tab:A.6a}
{
\fontsize{9.7}{9}\selectfont   %
\setlength{\tabcolsep}{6pt}
\renewcommand{\arraystretch}{1.3}
\begin{tabular}{lcccc}
\hline
\textbf{Approach} & \multicolumn{2}{c}{\textbf{Unmet power}} & \multicolumn{2}{c}{\textbf{Electricity price}} \\
\cmidrule(lr){2-3} \cmidrule(lr){4-5} 
 & RMSE & Trend Acc. & RMSE & Trend Acc. \\
\hline
CaReTS1 & 0.0650 $\pm$ 0.0014 & \textbf{0.9193 $\pm$ 0.0024} & 0.0418 $\pm$ 0.0011 & \textbf{0.9091 $\pm$ 0.0022} \\
CaReTS2 & \textbf{0.0646 $\pm$ 0.0016} & \textbf{0.9208 $\pm$ 0.0026} & \textbf{0.0408 $\pm$ 0.0013} & \textbf{0.9086 $\pm$ 0.0021} \\
CaReTS3 & \textbf{0.0641 $\pm$ 0.0021} & \textbf{0.9207 $\pm$ 0.0031} & 0.0422 $\pm$ 0.0013 & \textbf{0.9096 $\pm$ 0.0022} \\
CaReTS4 & 0.0654 $\pm$ 0.0023 & 0.9186 $\pm$ 0.0033 & \textbf{0.0412 $\pm$ 0.0013} & 0.9060 $\pm$ 0.0037 \\
Autoformer & 0.0683 $\pm$ 0.0020 & 0.8875 $\pm$ 0.0055 & 0.0437 $\pm$ 0.0018 & 0.8815 $\pm$ 0.0086 \\
FEDformer & 0.0841 $\pm$ 0.0003 & 0.8407 $\pm$ 0.0038 & 0.0843 $\pm$ 0.0006 & 0.7536 $\pm$ 0.0072 \\
Non-stationary & 0.1408 $\pm$ 0.0018 & 0.7341 $\pm$ 0.0126 & 0.1032 $\pm$ 0.0023 & 0.6842 $\pm$ 0.0157 \\
D-CNN-LSTM & 0.0651 $\pm$ 0.0018 & 0.8813 $\pm$ 0.0040 & 0.0503 $\pm$ 0.0015 & 0.7899 $\pm$ 0.0036 \\
TimesNet & 0.0648 $\pm$ 0.0014 & 0.8962 $\pm$ 0.0067 & 0.0434 $\pm$ 0.0021 & 0.8834 $\pm$ 0.0093 \\
Dlinear & 0.0744 $\pm$ 0.0004 & 0.8391 $\pm$ 0.0039 & 0.0664 $\pm$ 0.0006 & 0.7336 $\pm$ 0.0170 \\
Nlinear & 0.1105 $\pm$ 0.0002 & 0.8116 $\pm$ 0.0029 & 0.0903 $\pm$ 0.0005 & 0.7272 $\pm$ 0.0111 \\
TimeXer & \textbf{0.0637 $\pm$ 0.0016} & 0.9066 $\pm$ 0.0034 & \textbf{0.0417 $\pm$ 0.0015} & 0.8993 $\pm$ 0.0064 \\
TimeMixer & 0.1248 $\pm$ 0.0005 & 0.6646 $\pm$ 0.0037 & 0.0971 $\pm$ 0.0014 & 0.5687 $\pm$ 0.0091 \\
SOIT2FNN-MO & 0.1519 $\pm$ 0.0020 & 0.6955 $\pm$ 0.0027 & 0.1287 $\pm$ 0.0022 & 0.5946 $\pm$ 0.0039 \\
\hline
\end{tabular}
}
\end{table*}



\begin{table*}[t]
\centering
\caption{Comparison of SOTA algorithms for 15-8 multi-step forecasting}
\label{tab:A.6b}
{
\fontsize{9.7}{9}\selectfont   %
\setlength{\tabcolsep}{6pt}
\renewcommand{\arraystretch}{1.3}

\begin{tabular}{lcccc}
\hline
\textbf{Approach} & \multicolumn{2}{c}{\textbf{Unmet power}} & \multicolumn{2}{c}{\textbf{Electricity price}} \\
\cmidrule(lr){2-3} \cmidrule(lr){4-5} 
 & RMSE & Trend Acc. & RMSE & Trend Acc. \\
\hline
CaReTS1 & \textbf{0.0763 $\pm$ 0.0021} & \textbf{0.9095 $\pm$ 0.0030} & 0.0521 $\pm$ 0.0024 & 0.9090 $\pm$ 0.0019 \\
CaReTS2 & \textbf{0.0758 $\pm$ 0.0025} & \textbf{0.9085 $\pm$ 0.0040} & \textbf{0.0512 $\pm$ 0.0019} & \textbf{0.9183 $\pm$ 0.0033} \\
CaReTS3 & 0.0764 $\pm$ 0.0024 & 0.9066 $\pm$ 0.0025 & \textbf{0.0520 $\pm$ 0.0015} & \textbf{0.9107 $\pm$ 0.0023} \\
CaReTS4 & \textbf{0.0756 $\pm$ 0.0029} & \textbf{0.9090 $\pm$ 0.0050} & \textbf{0.0518 $\pm$ 0.0013} & \textbf{0.9185 $\pm$ 0.0011} \\
Autoformer & 0.0785 $\pm$ 0.0019 & 0.8856 $\pm$ 0.0031 & 0.0579 $\pm$ 0.0026 & 0.8608 $\pm$ 0.0082 \\
FEDformer & 0.1097 $\pm$ 0.0004 & 0.8165 $\pm$ 0.0028 & 0.1035 $\pm$ 0.0014 & 0.7607 $\pm$ 0.0071 \\
Non-stationary & 0.1581 $\pm$ 0.0018 & 0.7250 $\pm$ 0.0031 & 0.1286 $\pm$ 0.0015 & 0.7118 $\pm$ 0.0099 \\
D-CNN-LSTM & 0.0790 $\pm$ 0.0024 & 0.8663 $\pm$ 0.0027 & 0.0631 $\pm$ 0.0011 & 0.8007 $\pm$ 0.0026 \\
TimesNet & 0.0776 $\pm$ 0.0011 & 0.8939 $\pm$ 0.0017 & 0.0578 $\pm$ 0.0038 & 0.8827 $\pm$ 0.0101 \\
Dlinear & 0.0805 $\pm$ 0.0003 & 0.8179 $\pm$ 0.0015 & 0.0633 $\pm$ 0.0004 & 0.7492 $\pm$ 0.0038 \\
Nlinear & 0.1360 $\pm$ 0.0002 & 0.7855 $\pm$ 0.0017 & 0.1127 $\pm$ 0.0006 & 0.7424 $\pm$ 0.0025 \\
TimeXer & 0.0769 $\pm$ 0.0033 & 0.8934 $\pm$ 0.0060 & 0.0532 $\pm$ 0.0034 & 0.9049 $\pm$ 0.0042 \\
TimeMixer & 0.1307 $\pm$ 0.0003 & 0.7125 $\pm$ 0.0011 & 0.1240 $\pm$ 0.0003 & 0.6252 $\pm$ 0.0051 \\
SOIT2FNN-MO & 0.1689 $\pm$ 0.0026 & 0.6886 $\pm$ 0.0023 & 0.1304 $\pm$ 0.0019 & 0.6599 $\pm$ 0.0039 \\
\hline
\end{tabular}
}
\end{table*}

\section{Conclusion}\label{sec:5}

Short-term energy forecasting poses inherent challenges in simultaneously achieving high numerical accuracy and providing structured outputs that reflect short-term directional movements. To address this challenge, this paper presents a trend-aware multi-task learning framework that jointly models directional movements and deviation magnitudes, enabling forecasting outputs that are both numerically accurate and operationally interpretable. An uncertainty-aware task weighting strategy is further incorporated to adaptively balance heterogeneous learning objectives and stabilize joint optimization. Based on this framework, four model variants (CaReTS1 -- CaReTS4) are designed to be compatible with mainstream encoders, including CNN, LSTM, and Transformer. Experiments on real-world energy forecasting tasks demonstrate that the proposed approach achieved competitive numerical accuracy and consistently strong trend prediction accuracy. These results are particularly valuable for short-term energy system operation, where reliable directional signals can support practical tasks such as electricity market bidding, short-term energy dispatch adjustment, and risk-aware operational planning. Future work will investigate the extension of the proposed framework to longer forecasting horizons and its integration with uncertainty-aware operational optimization in large-scale energy systems.


\section*{Acknowledgments}
This work was supported in part by the Clean Energy and Equitable Transportation Solutions (CLEETS) NSF–UKRI Global Center under NSF Award No. 2330565 and UKRI Award No. EP/Y026233/1.


\bibliographystyle{elsarticle-num} 
\bibliography{CaReTs}


\end{document}